\documentclass{article}

\usepackage{PRIMEarxiv}
\usepackage{adjustbox}
\usepackage[normalem]{ulem}
\useunder{\uline}{\ul}{}
\usepackage{caption}
\usepackage{subcaption}
\usepackage{multirow}

\usepackage[table,xcdraw]{xcolor}
\usepackage[utf8]{inputenc} 
\usepackage[T1]{fontenc}    
\usepackage{hyperref}       
\usepackage{url}            
\usepackage{booktabs}       
\usepackage{amsfonts}       
\usepackage{nicefrac}       
\usepackage{microtype}      
\usepackage{lipsum}
\usepackage{fancyhdr}       
\usepackage{tikz}
\usepackage{comment}
\usepackage{amsmath,amssymb} 
\usepackage{color}
\usepackage{graphicx}       
\graphicspath{{media/}}     

\title{Spiking Neural Networks for Frame-based and Event-based Single Object Localization} 

\author{
  Sami Barchid$^1$, José Mennesson$^{1,2}$, Jason Eshraghian$^{3,4}$, Chaabane Djéraba$^{1}$, Mohammed Bennamoun$^{4}$ \\
  $^1$Univ. Lille, CNRS, Centrale Lille, UMR 9189 CRIStAL, F-59000 Lille, France \\
  $^2$IMT Nord Europe, Institut Mines-Télécom, Centre for Digital Systems, F-59000 Lille, France \\
  $^3$ Department of Electrical Engineering and Computer Science, University of Michigan, Ann Arbor, MI 48105, USA\\
  $^4$ The University of Western Australia, Perth, WA 6009, Australia \\
  \texttt{ \{sami.barchid, chabane.djeraba\}@univ-lille.fr} \\ \texttt{jose.mennesson@imt-nord-europe.fr}\\
  \texttt{\{jason.eshraghian, mohammed.bennamoun\}@uwa.edu.au}\\
}

\begin{document}
\maketitle

\begin{abstract}
Spiking neural networks have shown much promise as an energy-efficient alternative to artificial neural networks. However, understanding the impacts of sensor noises and input encodings on the network activity and performance remains difficult with common neuromorphic vision baselines like classification. Therefore, we propose a spiking neural network approach for single object localization trained using surrogate gradient descent, for frame- and event-based sensors. We compare our method with similar artificial neural networks and show that our model has competitive/better performance in accuracy, robustness against various corruptions, and has lower energy consumption. Moreover, we study the impact of neural coding schemes for static images in accuracy, robustness, and energy efficiency. Our observations differ importantly from previous studies on bio-plausible learning rules, which helps in the design of surrogate gradient trained architectures, and offers insight to design priorities in future neuromorphic technologies in terms of noise characteristics and data encoding methods.
\end{abstract}


\section{Introduction}
\label{sec:introduction}

Often referred to as the third generation of neural networks \cite{maass}, Spiking Neural Networks (SNNs) that are derived from models of biological neurons have shown the potential for more energy-efficient and bio-plausible artificial intelligence (AI) when compared to traditional Artificial Neural Networks (ANNs) \cite{truenorth,azghadi2020hardware,loihi,furber2014spinnaker}. 

In computer vision, SNNs have demonstrated great progress in image classification, but the domain remains relatively underexplored beyond classification baselines \cite{dvsgesture,nmnist}. 
The emergence of bio-inspired vision sensors, such as dynamic vision sensor (DVS) cameras \cite{dvs_survey}, has paved the way for completely neuromorphic vision solutions by coupling them together with SNNs. However, the encoding mechanism at the input has a direct impact on the degree of network activity, and thus, the overall power efficiency of the network \cite{survey_snn}. Furthermore, event-based image sensors are susceptible to electron and photon noise, and vast efforts have gone into reducing the impact both at the photoreceptor pixel level, and in digital post-processing \cite{graca2021unraveling,lenz_gregor_2021_5079802}. Understanding the effect of various noise types on different neural encoding schemes is important for determining how much noise mitigation should be accounted for at the pixel level.



\begin{figure}[t]
\centering
\includegraphics[width=	0.80\textwidth]{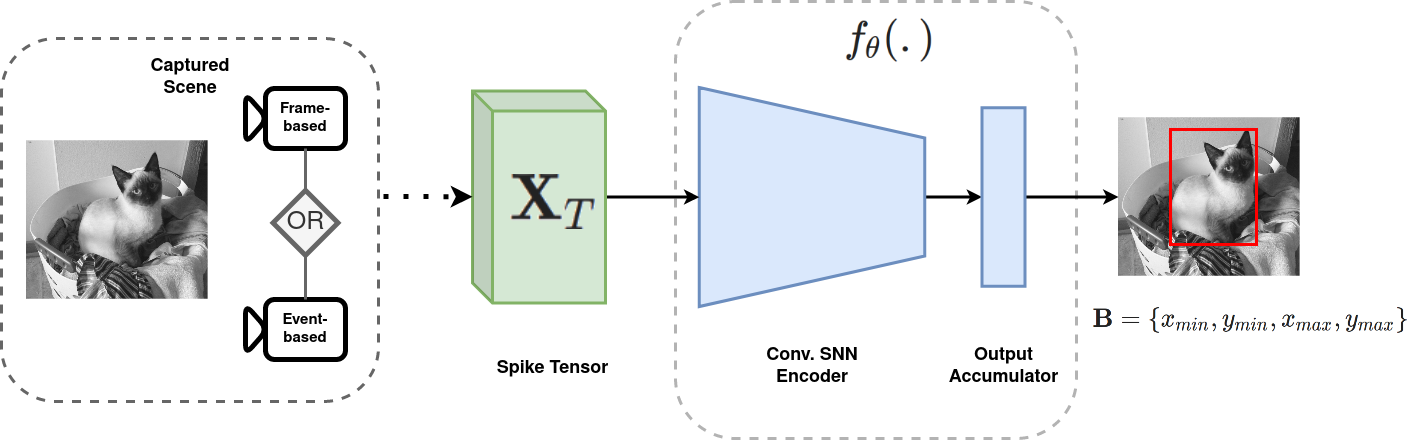}
\caption{Overview of the proposed method based on a convolutional spiking neural network. Our approach can adapt to frame-based or event-based inputs}
\label{fig:overview}
\end{figure}

In this paper, we provide a rigorous empirical evaluation of bio-inspired single object localization that accounts for a broad sweep of neural encoding schemes, noise perturbation analyses, and overall energy efficiency.
Our presented approach using the backpropagation-trained convolutional SNN architecture in Fig.~\ref{fig:overview} outperforms similar ANN architectures in several cases and is more robust to common image corruptions (See Section \ref{sec:experiments}). We investigate various popular neural coding schemes for static images to determine the advantages and drawbacks in terms of energy efficiency, inference latency, and robustness. In doing so, our results provide insight to how spike-based sensing may be expanded beyond responding only to local changes \cite{dvs_survey}.

The contributions of this paper are summarized as follows:
\begin{enumerate}
\item We propose a novel approach for single object localization using a directly trained convolutional SNN. An output accumulator module on top of the SNN is designed to obtain precise bounding box predictions from the binary output spikes. In our experiments, we show that our method can effectively adapt to different kinds of inputs during training.
\item We study the performance of our approach on both static images with the Oxford-IIIT-Pet dataset \cite{Oxford-IIIT-Pet} and event-based inputs on N-Caltech101 \cite{nmnist}. We find that our method has similar or better performance (in accuracy) than a similar ANN architecture and does not need a large number of inference time-steps like other works on SNNs \cite{large_t}. Our model also proves to be more robust to noise depending on its type and on the experienced coding scheme. Finally, we observe better energy efficiency for our approach, with orders of magnitude lower energy consumption than similar ANNs.
\item We summarize our observations on various neural coding schemes for SNNs trained with surrogate gradient descent on static images according to our experiments on accuracy, inference efficiency, robustness, and energy consumption. We find that our conclusions differ from previous studies on shallower SNNs trained with STDP learning rules \cite{falez_thesis,neural_coding}.
\end{enumerate}

\section{Related Works}
\label{sec:related_works}

Most works designed to train SNNs can be categorized depending on their learning paradigm. In this section, we discuss three of the main learning strategies and their ability to address modern computer vision problems (object detection, segmentation, etc.).

\subsection{Unsupervised Bio-plausible Learning} Many works try to exploit bio-plausible learning rules to train SNNs for various machine learning tasks. This strategy is mostly based on Hebbian-like \cite{hebbian} or STDP \cite{stdp} learning rules. Even if this paradigm is strongly inspired by observed biological mechanisms and is easy to implement on neuromorphic hardware, it remains limited to small neural networks \cite{falez_thesis} and deals with basic computer vision tasks (e.g., digit recognition \cite{stdp_nmnist}, low-level feature extraction \cite{mireille}). Typically, models trained with STDP are limited to 2 or 3 layers, which limits the complexity of extracted features. Some works \cite{falez_threshold} try to adapt these rules to train deeper networks but are still far from achieving the same level of performance of ANNs \cite{how_far_are_we}.

\subsection{ANN to SNN Conversion} Instead of direct training, some works focus on converting trained ANNs into SNNs so that the deployed solution can benefit from both the energy efficiency of SNNs and the high performance of ANNs \cite{ann2snn_3,ann2snn_1,ann2snn_2}. Many works already achieve similar performance to ANNs \cite{sota_ann2snn}, and there exist approaches for common computer vision tasks such as object detection \cite{spiking_yolo,snn_lidar} or semantic segmentation \cite{snn_segmentation}. However, these methods rely on rate-coded inputs, which results in a high latency \cite{ann2snn_4} and intense energy consumption (sometimes higher than ANNs), even on neuromorphic hardware \cite{loihi_survey}. Because of rate coding, this strategy is not suited for event-based sensors or other neural coding schemes that can be more efficient than rate coding \cite{neural_coding}. Also due to rate coding, converted SNNs cannot benefit from their natural ability to process temporal information.

On the other hand, \cite{spiking_yolo} already deals with a similar regression problem as ours (i.e. object detection), and achieves satisfactory numerical precision from binary spike trains. However, their approach requires intensive rate coding of static images with a large number of time-steps, which can lead to poor efficiency. On the contrary, our method works with various neural coding schemes and achieves good numerical precision without requiring a large number of time-steps, wich makes it more efficient and versatile. 

\subsection{Supervised Learning} Early works \cite{tempotron,resume} focus on supervised learning for single-layered SNNs. Subsequent works \cite{spikeprop,span} try to adapt the backpropagation algorithm to train multi-layered and thus more complex networks. Recently, we have seen the emergence of surrogate gradient learning methods \cite{snntorch,surrogate,surrogate_3,STBP,surrogate_2}, which achieve similar performance to ANNs and have been broadly adopted to explore various common computer vision tasks \cite{snn_segmentation,stereospike}. Recently, on-chip implementations of native backpropagation have become more prevalent \cite{frenkel2022,renner2021backpropagation}, and
enable the design of deep SNNs that can deal with various types of time-varying inputs (not only rate coding in most ANN to SNN conversion approaches) \cite{STBP}. In addition, this strategy does not require a large number of inference time-steps \cite{plif}, which makes them suitable for energy-efficient algorithms. Consequently, surrogate gradient learning has shown much success in the development of SNN-based solutions directly trained in the spike domain. In computer vision, surrogate gradient learning enables the training of deeper architectures \cite{sew_resnet,large_t}, and some popular tasks have already been explored (depth prediction \cite{stereospike}, semantic segmentation \cite{snn_segmentation}). However, there are still a large number of vision tasks that remain unexplored, which induce a lack of available solutions in state-of-the-art to develop neuromorphic applications, notably when an object instance must be spatially detected. On that basis, we adopt this strategy to propose an SNN method capable of addressing such challenge.
\section{Deep Learning for SNNs}
\label{sec:background}
In this section, we briefly discuss the spiking neuron model employed in our approach. In addition, we give a short description of the training method to directly train our deep SNN-based model.

\subsection{Integrate-and-Fire (IF) Neurons}
\label{subsec:neuron_model}
\begin{figure}[t]
\centering
\includegraphics[width=	0.33\textwidth]{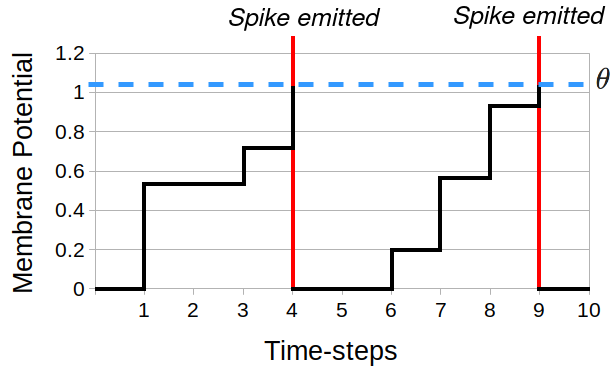}
\caption{The membrane potential of an IF neuron. A spike (shown in red) is emitted when its membrane potential exceeds a certain threshold $\theta$, and the membrane potential returns to its resting value}
\label{fig:ifneuron}
\end{figure}

Spiking neuron models aim to describe the dynamics of biological neurons. They vary in terms of complexity and bio-plausibility \cite{survey_neuron_models}. In this paper, we use one of the most popular neuron models: the IF neuron \cite{if_neuron}. It accumulates input spikes weighted by synaptic weights by increasing the hidden state (known as the ``membrane potential'') and emits a spike when a threshold is exceeded. After spike emission, the membrane potential is reset to its resting value (defined as 0 in our work). Figure \ref{fig:ifneuron} illustrates the evolution of the membrane potential of an IF neuron.
\begin{align}
\label{equation_if_neuron}
V_t^l & = V_{t-1}^l + W^{l-1} S^{l-1}_{t-1} - \theta S_t \\
S_t^l & = \Theta(V_t^l - \theta) 
\end{align} Equation \ref{equation_if_neuron} describes the discretized dynamics of a layer $l$ of IF neurons. $V_t^l$ denotes the membrane potentials from neurons of layer $l$ at a certain time-step $t$, $W^{l-1}$ denotes the pre-synaptic weights from the preceding layer $l-1$, and $S^{l-1}_{t-1}$ denotes the outputs from the pre-synaptic neurons. The output $S^l_{t}$ is composed of $1$'s when a neuron's membrane potential exceeds its threshold $\theta$, and $0$'s otherwise. This behavior corresponds to the Heaviside step function $\Theta(\cdot)$. Finally, the rightmost term of Equation \ref{equation_if_neuron} defines the resting mechanism after a spike.

\subsection{Surrogate Gradient Learning}
A common and effective method to train deep SNNs is to express them as a recurrent neural network and use backpropagation through time \cite{bptt} to adapt the synaptic weights \cite{STBP}. However, the derivative of $\Theta(\cdot)$ is 0 almost everywhere, but when there is a spike, which breaks the gradient chain and prevents the SNN from learning effectively. 
Surrogate gradient learning \cite{surrogate_3,surrogate_2} aims to address this problem by using the derivative of a continuous surrogate function $\sigma(\cdot)$ during backpropagation instead of the derivative of $\Theta(\cdot)$. In our method, we define $\sigma(x) = \frac{1}{\pi} \arctan(\frac{\pi}{2}\alpha x) + \frac{1}{2}$, with $\alpha = 2$. Consequently, a multi-layer SNN can be directly trained despite the non-differentiability problem.


\section{Methodology}
\label{sec:methodology}

\subsection{Problem Formulation}
\label{subsec:problem_formulation}

Given a stream of input spikes obtained from an event-based sensor or a static image (using a neural coding scheme), the objective is to predict the bounding box coordinates $\mathbf{B} = \{x_{min}, y_{min}, x_{max}, y_{max}\}$, where $(x_{min}, y_{min})$ and $(x_{max}, y_{max})$ are the upper-left and bottom-right corners of the bounding box, respectively. To do so, we design an SNN-based model $f_\theta(.)$ with $\theta$ representing the set of trainable parameters (synaptic weights and biases) such that :
\begin{align}
f_{\theta}(\mathbf{X}_T) = \{x_{min}, y_{min}, x_{max}, y_{max}\} = \mathbf{B}
\end{align}
where $\mathbf{X}_T \in \{0,1\}^{T \times C \times H \times W} = \{ X_t\}_{t=1}^T$  is the binary tensor representing the stream of input spikes discretized into $T$ time-steps, obtained from a frame-based or event-based sensor with a resolution of $H \times W$, and $C$ channels (e.g., $C = 3$ for an RGB camera, $C = 2$ for a DVS camera, etc.).

\subsection{Model Architecture}
\label{subsec:network_architecture}

\begin{figure}[t]
\centering
\includegraphics[width=0.70\textwidth]{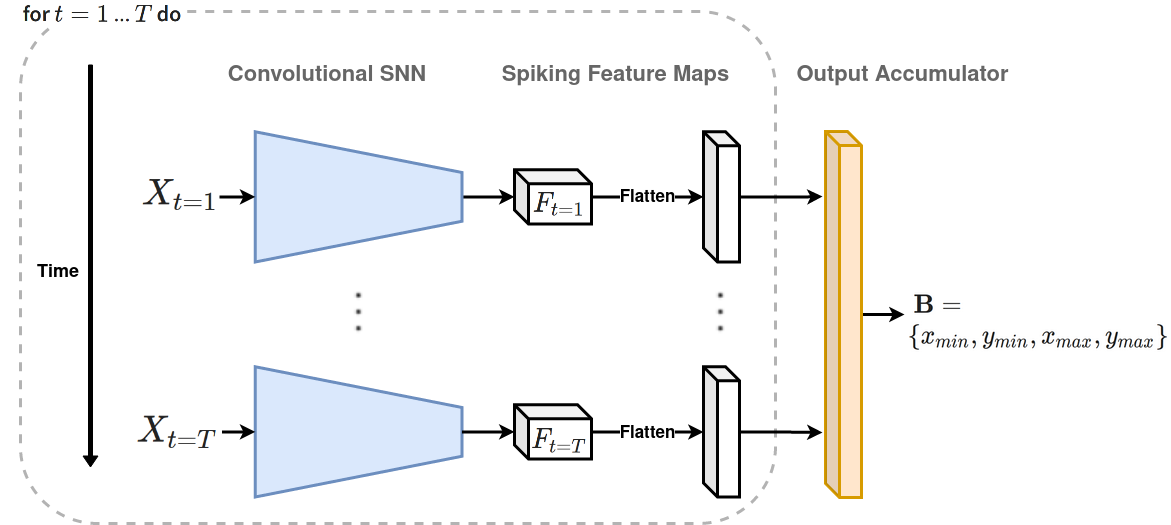}
\caption{The proposed model architecture based on a Convolutional SNN}
\label{fig:architecture}
\end{figure}

The proposed model consists of two modules:  a convolutional SNN encoder composed of several convolutional IF neuron layers (e.g., a SEW-ResNet \cite{sew_resnet} model) and an output accumulator module. Figure \ref{fig:architecture} illustrates the proposed architecture. The SNN encoder extracts spiking feature maps $\mathbf{F}_T = \{ F_t\}_{t=1}^{T}$ from the input spikes $\mathbf{X}_T$. These feature maps are flattened and fed into the output accumulator module, with the purpose of obtaining the final bounding box prediction $\mathbf{B}$ from the spatio-temporal feature maps.

Since object localization is a regression problem, the main challenge related to the output accumulator is to convert these spiking feature maps into numerical values which represent precise bounding boxes coordinates. In addition, the output accumulator should adapt to different types of inputs during training (e.g., rate coded images, event-based streams, etc.). Inspired from \cite{snn_segmentation}, our output accumulator module is a fully connected layer of IF neurons with an infinite threshold, which means that the membrane potentials of this last layer accumulate spikes from each time-step. At the end of the $T$ time-steps, the membrane potential of these neurons determines the bounding box prediction $\mathbf{B}$. In fact, the output accumulator is an intuitive design that allows us to focus on how the SNN encoder processes visual inputs (from both event- and frame-based sensors) without being influenced by more complex modules that could produce biases in our empirical analysis.

The whole model is trained end-to-end using the Distance-IoU loss \cite{diou}.

\subsection{Input Representation}
\label{subsec:input_representations}
As shown on the left of Figure \ref{fig:overview}, the input tensor $\mathbf{X}_T$ is obtained either from an event-based sensor, or a static image using a neural coding scheme.  In this section, we introduce the strategies used to obtain this spike tensor for both modalities.

\subsubsection{Event-based Inputs}
For event-based sensors, events are accumulated into a fixed number of $T$ event frames, sliced along the time axis of the whole sequence. 

\subsubsection{Static Images through Neural Coding Scheme}
Static images are commonly converted into spike trains using a specific process known as the \textit{``neural coding scheme''} \cite{neural_coding}. In this work, we investigate various popular schemes. \textbf{Rate Coding:} each input pixel is considered as the probability a spike occurs at each time-step, which results in a frequency of spikes proportional to the original value. \textbf{Time-To-First-Spike (TTFS) Coding:} information is encoded through precise spike timings. Each pixel value fires at most one spike, where a high pixel value results in an early spike and a low pixel value produces a late or no spike. \textbf{Phase Coding \cite{phase_coding}:} the integer intensity of each pixel (i.e., from 0 to 255) is converted into an 8-bit representation (i.e., a set of 0s and 1s). The "1" signals are generated usong an 8-phase cycle that is repeated until $T$ time-steps are created. To replicate the significance of each bit in their binary representation, spikes are weighted according to their related phase, given by $w_s(t) = 2^{-(1+mod(t-1, 8))}$. \textbf{Saccades Coding:}  we simulate the same process as previous works on event-based cameras \cite{cifar10_dvs,nmnist} by translating the image in three saccades and applying a delta modulation process \cite{snntorch}, which emits a spike when the change of intensity between two frames for the same pixel exceeds a certain threshold. \textbf{Trainable Coding:} the first convolutional layer of our model directly takes the input image to obtain low-level spiking feature maps. These feature maps are then repeated over the $T$ time-steps, which makes the first convolutional layer an encoder that can be trained directly to obtain an optimal coding scheme.
\section{Experiments}
\label{sec:experiments}

In addition to the baseline localization accuracy, our experiments aim to evaluate the performance of the proposed approach on three aspects: inference efficiency, energy consumption, and robustness on several image corruptions. Moreover, our SNN-based approach is compared to a similar ANN architecture. Nevertheless, a comparison with state-of-the-art (and generally deeper and more complex) ANNs is not provided, since it could lead to biased conclusions on the differences between ANNs and SNNs on the studied aspects for localization tasks. On the other hand, we study the impact of various neural coding schemes for frame-based inputs following these three aspects.

As for the ANN baseline, static images are directly fed into the model. For event-based inputs, each of the $T$ event-frames is fed into the convolutional layers. The $T$ resulting feature maps are summed before passing through fully connected layers to obtain the prediction.

\subsection{Datasets and Metrics}
\label{subsec:dataset_and_metrics}

The Oxford-IIIT-Pet dataset \cite{Oxford-IIIT-Pet} is used to evaluate the performance on static images. It consists of images containing strictly one cat or one dog in a complex (and thus challenging) environment. The dataset is split into 6000 samples for the training set and 1300 images for the validation set.

To evaluate our method with event-based inputs, we use a subset of N-Caltech101 \cite{nmnist}. N-Caltech101 is the spiking version of the frame-based Caltech101 dataset \cite{caltech101}. This event-based dataset is obtained following the same strategy proposed in \cite{nmnist}, i.e., an event-based camera captures each image from a screen while it is mounted on a motorized pan-tilt that mimics saccadic eye movements. Our subset is composed of 2035 training samples and 609 samples for validation. The samples were selected from the classes containing $\ge 100$ samples, in order to limit data imbalance.

Since our approach is aimed at localizing strictly one object per sample, we measure the performance using the mean intersection over union ($mIoU$).

\subsection{Implementation Details}
\label{subsec:implementation_details}
The experiments are conducted using the SpikingJelly 0.0.0.8 framework \cite{spikingjelly}, an SNN simulator that runs on PyTorch \cite{pytorch}. All experiments are conducted using an NVIDIA 2080Ti GPU. We use a SEW-Resnet-18 \cite{sew_resnet} architecture for the convolutional SNN encoder. For comparison with a similar ANN model, a ResNet-18 \cite{resnet} architecture is employed. All models are trained for 150 epochs using the Adam optimizer \cite{adam_optimizer}. The learning rate for each architecture is found using a learning rate finder \cite{lr_finder}. Every sample is resized to a resolution of $224 \times 224$. To ensure the comparison with previous works on single object localization with SNNs \cite{sami_decolle}, static images are converted to grayscale ($C = 1$). As for the event-based inputs, we use the common On/Off polarity channels ($C = 2$) of DVS cameras.

\subsection{Analysis on Time-Steps Inference}
\label{subsec:analysis_on_timesteps}

\begin{figure}[b]
     \centering
     \begin{subfigure}[b]{0.45\textwidth}
         \centering
         \includegraphics[width=\textwidth]{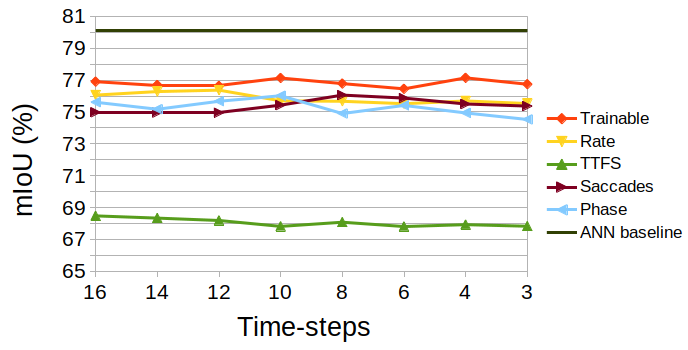}
         \caption{Oxford-IIIT-Pet}
         \label{fig:static_inference}
     \end{subfigure}
     \hfill
     \begin{subfigure}[b]{0.40\textwidth}
         \centering
         \includegraphics[width=\textwidth]{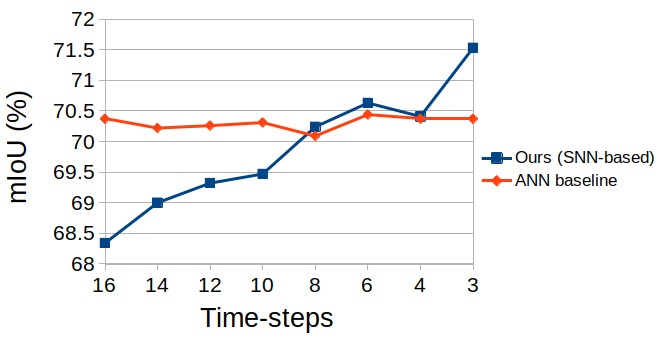}
         \caption{Subset of N-Caltech101}
         \label{fig:dvs_inference}
     \end{subfigure}
        \caption{Performance of our approach vs. $T$, the total number of time-steps for inference}
        \label{fig:inference_results}
\end{figure}

Previous works \cite{large_t} on backprop-trained SNNs for computer vision show that better performance should be expected with larger $T$ time-steps, as it allows more spikes to be integrated by the SNN at the cost of efficiency. On the other hand, more recent works \cite{plif} show the opposite, with better accuracy and a reduced number $T$ of time-steps. However, the importance of larger $T$ value is still debatable because the SNN models of previous works was never similar, making it difficult to draw a definitive conclusion. In this section, we answer this question by using the same model and studying its performance with different values of $T$.

Figure \ref{fig:inference_results} shows the performance of our approach for both event-based and frame-based inputs depending on the total number of time-steps $T$. For both contexts, we can see that our SNN-based method has consistent results across all coding schemes and with event-based inputs, which shows the ability of our output accumulator module to adapt to various types of inputs. In addition, our model has similar performance compared to its ANN counterpart.  Specifically, our method has slightly worse performance for static images but outperforms the ANN on event-based inputs.

As for the comparison between neural coding schemes, our results show that there is no strong correlation between $T$ and the final performance. Consequently, unlike previous works \cite{large_t}, our method does not need a large number of time-steps to perform consistently. We observe different results from other studies on coding schemes using bio-plausible unsupervised learning rules, such as STDP \cite{neural_coding}. While TTFS coding is considered better in terms of accuracy using STDP (e.g., compared to rate coding), TTFS instead leads to lower results than any other scheme. This can be explained by the fact that STDP learning rules may not be adapted to spike-intensive inputs from rate-coded images, contrary to surrogate gradient learning. Other coding schemes have similar performance, with a slight advantage for the trainable coding scheme.

In the context of event-based sensors, we observe better results with smaller values of $T$, contrary to the baseline ANN that shows similar results across all values. Consequently, our method outperforms the ANN baseline for $T \leq 10$. We report the best performance of $71.53$ \% $mIoU$ for $T = 3$. In fact, saccadic motion is broadly accepted as a biologically plausible encoding scheme. However, when applied to static images, there is little to no temporally relevant information within each sample of data \cite{jason_suggestion}. As a result, when the sensor is static, no useful information is being passed and so integrating the full temporal window of information within a shorter $T$ avoids sequence steps which are sparse and do not provide useful information to the network.

\begin{table}[t]
\centering
\caption{Best performance (based on $T$) for each coding scheme}
\label{tab:best_results}
\begin{tabular}{c|ccccc|c|c}
\textbf{}    & \textbf{Learnable} & \textbf{Rate} & \textbf{TTFS} & \textbf{Phase} & \textbf{Saccades} & \textbf{\begin{tabular}[c]{@{}c@{}}ANN\\ baseline\end{tabular}} & \textbf{\begin{tabular}[c]{@{}c@{}}\cite{sami_decolle}\\ (rate coding)\end{tabular}} \\ \hline
$mIoU$ (\%)  & \uline{77.14}     & 76.37         & 68.48         & 76.02          & 76.06             & \textbf{80.11}                                                           & 63.2                                                                 \\
\textbf{$T$} & 4                  & 12            & 16            & 10             & 8                 & -                                                               & 1000                                                                
\end{tabular}
\end{table}

Table \ref{tab:best_results} shows the best-performing models for each coding scheme, with the related value of $T$. All versions of our model outperforms previous work \cite{sami_decolle} by a large margin in both $mIoU$ and inference efficiency. They have slightly worse results than the ANN baseline (from 11.63 to 2.97 \%, depending on the coding scheme).

\subsection{Analysis on Corruption Robustness}
\label{subsec:analysis_on_robustness}

Similarly to \cite{benchmark_corruptions}, we investigate the robustness of our method using several types of common image corruptions, and with a growing level of severity from $1$ (weakest) to $5$ (strongest). Models trained with $T = 8$ are used for all the experiments of this section.

For a specific corruption $corr$ with a given severity level $sev$, the drop of performance of our model is given by the “Relative Accuracy Drop" \cite{snn_segmentation} $RAD^{corr}_{sev}$ such that:

 \begin{equation}
 RAD^{corr}_{sev} = \frac{mIoU_{clean} - mIoU^{cor}_{sev}}{mIoU_{clean}} \times 100
 \end{equation}

where $mIoU_{clean}$ and $mIoU^{corr}_{sev}$ are the mIoU metrics without and with the defined corruption, respectively.

To estimate the overall robustness of our method against a specific corruption, we also introduce the “Mean Relative Accuracy Drop" ($mRAD^{corr}$) :
\begin{align}
mRAD^{corr} = \frac{1}{5} \times \sum\limits_{sev = 1}^{5} RAD^{corr}_{sev}
\end{align}

\subsubsection{Corruptions for Frame-based Sensors}
\label{subsubsec:studied_corruptions}

The following corruptions of static images are evaluated. \textit{Gaussian Noise:} image noise that happens in low-lighting environments. \textit{Salt \& Pepper Noise:} image degradation that shows sparsely “broken" pixels (i.e., only white or black). \textit{JPEG Compression:} degradation due to a lossy jpeg compression of the image. \textit{Defocus Blur:} specific blur effect occurring when the camera is out of focus. \textit{Frost Perturbation:} frost occlusions that occur when a camera lens is covered by ice crystals.

\subsubsection{Corruptions for Event-based Sensors}
With event-based sensors, two noises that represent known effects on DVS cameras are explored. \textit{Hot Pixels:} corruption that occurs due to broken pixels (i.e., always active) in the sensor. It consists of random pixels that fire at a very high rate \cite{v2e}. \textit{Background Activity:} noise that appears when the output of a pixel changes under constant illumination. This corruption can be effectively simulated by a time-independent Poisson noise \cite{background_activity}. It is worth mentioning that background activity noise already occurs in our event-based dataset since DVS cameras are very sensitive to it. Therefore, this corruption is aimed at increasing the background activity noise.
\begin{table}[t]
\centering
\caption{Mean relative accuracy drop comparison on static images. Scores better than the ANN baseline are highlighted in green}
\label{tab:mean_acc_drop_static}
\begin{adjustbox}{max width=0.99\textwidth}
\begin{tabular}{c|c|c|c|c|c}
                      & \textbf{Gaussian}               & \textbf{JPEG Compression}             & \textbf{Salt\&Pepper}                 & \textbf{Defocus Blur}                 & \textbf{Frost Perturbation}                                  \\ \hline
\textbf{ANN Baseline} & 4.35                                  & 0.3                                   & 4.56                                  & 4.05                                  & 4.61                                                         \\ \hline
\textbf{Trainable}    & 6.34                                  & \cellcolor[HTML]{9AFF99}0.22          & 6.62                                  & \cellcolor[HTML]{9AFF99}3.1           & 6.08                                                         \\
\textbf{Rate}         & \cellcolor[HTML]{9AFF99}0.87          & \cellcolor[HTML]{9AFF99}\textbf{0.04} & \cellcolor[HTML]{9AFF99}0.92          & \cellcolor[HTML]{9AFF99}0.87          & \cellcolor[HTML]{9AFF99}{\color[HTML]{333333} \textbf{3.39}} \\
\textbf{TTFS}         & \cellcolor[HTML]{9AFF99}\textbf{0.38} & \cellcolor[HTML]{9AFF99}0.09          & \cellcolor[HTML]{9AFF99}\textbf{0.15} & \cellcolor[HTML]{9AFF99}\textbf{0.48} & 23.57                                                        \\
\textbf{Saccades}     & 23.41                                 & 0.87                                  & 21.83                                 & 6.24                                  & \cellcolor[HTML]{9AFF99}{\color[HTML]{333333} 4.52}          \\
\textbf{Phase}        & 5.51                                  & 0.79                                  & 9.06                                  & \cellcolor[HTML]{9AFF99}2.13          & 7.39                                                        
\end{tabular}
\end{adjustbox}
\end{table}
\subsubsection{Results}
\label{subsubsec:robustness_results}
\begin{figure}[t]
     \centering
     \begin{subfigure}[b]{0.2\textwidth}
         \centering
         \includegraphics[width=\textwidth]{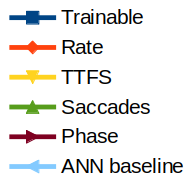}
         \caption{Legend}
         \label{fig:legend_frame_noise}
     \end{subfigure}
     \hfill
     \begin{subfigure}[b]{0.3\textwidth}
         \centering
         \includegraphics[width=\textwidth]{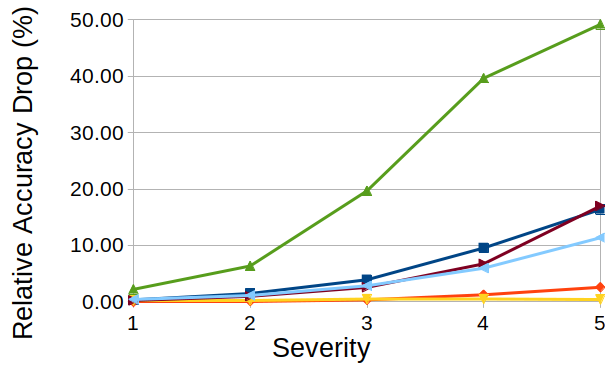}
         \caption{Gaussian Noise}
         \label{fig:gaussian_evolution}
     \end{subfigure}
     \hfill
     \begin{subfigure}[b]{0.3\textwidth}
         \centering
         \includegraphics[width=\textwidth]{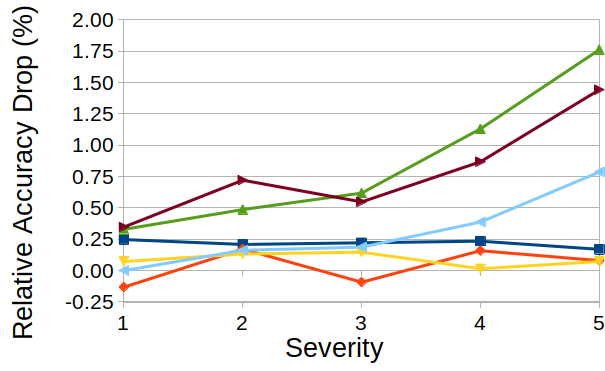}
         \caption{JPEG Compression}
         \label{fig:jpeg_evolution}
     \end{subfigure}
     \\
     \begin{subfigure}[b]{0.3\textwidth}
         \centering
         \includegraphics[width=\textwidth]{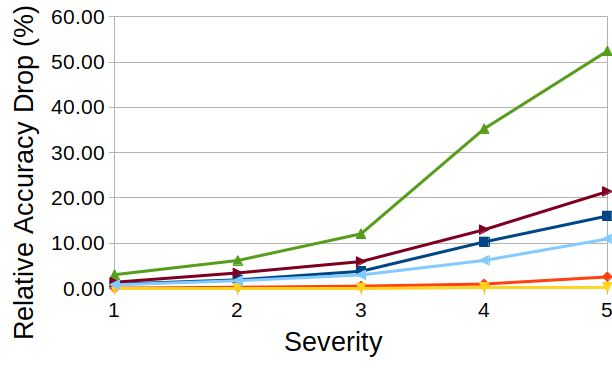}
         \caption{Salt and Pepper Noise}
         \label{fig:sp_evolution}
     \end{subfigure}
     \hfill
     \begin{subfigure}[b]{0.3\textwidth}
         \centering
         \includegraphics[width=\textwidth]{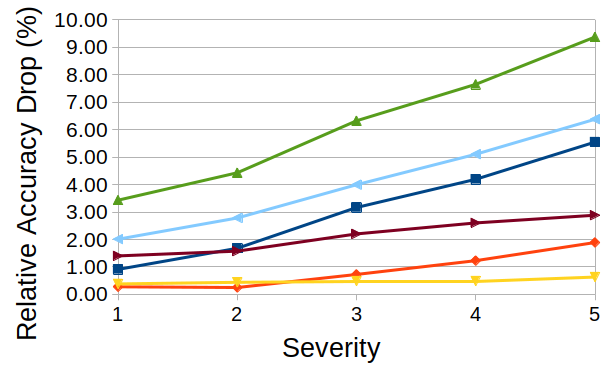}
         \caption{Defocus Blur}
         \label{fig:defocus_evolution}
     \end{subfigure}
     \hfill
     \begin{subfigure}[b]{0.3\textwidth}
         \centering
         \includegraphics[width=\textwidth]{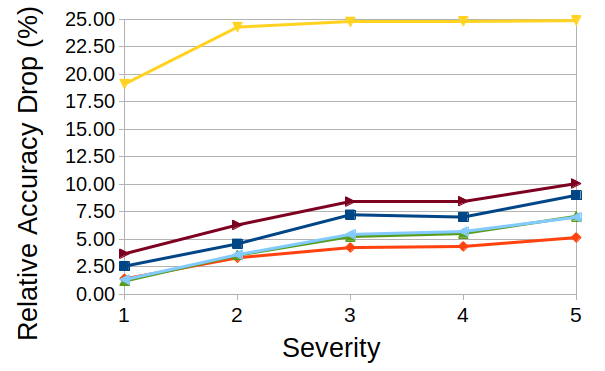}
         \caption{Frost Perturbation}
         \label{fig:frost_evolution}
     \end{subfigure}
        \caption{Evaluation of the relative accuracy drop score in terms of corruption severity level for each coding scheme}
        \label{fig:evolution_severity}
\end{figure}

We firstly investigate the overall robustness of each neural coding scheme, shown in Table \ref{tab:mean_acc_drop_static}. The evaluation of this robustness in terms of the increasing severity of the noise is shown in Figure \ref{fig:evolution_severity}. Our approach seems to be more robust than the ANN baseline on specific cases, i.e., using a certain neural coding to deal with a target noise. Unlike previous studies with STDP \cite{falez_thesis,neural_coding}, TTFS coding is much more robust than other studied schemes. It is more commonly thought that rate coding is more robust to TTFS as multiple spikes provide an opportunity for errors to be averaged out in rate codes \cite{snntorch,neural_coding}. Our results here offer an alternative theory, by showing that the relative accuracy drop is generally constant, even against corruptions with high severity. This may be explained by noises having only a marginal effect on the logarithmic dependency between input intensity and spike times, whereas rate codes frequent spiking provide additional opportunities for noise to manifest. As shown in Figure \ref{fig:frost_evolution}, a high sensitivity to frost perturbation can be noticed even with a low severity, which suggests that our model strongly relies on the early input spikes, which are highly corrupted by the bright pixels composing frost perturbations. On the other hand, saccades coding is the least robust scheme by a large margin. More precisely, it shows similar robustness to other schemes on low severity but seems to be not adapted to very noisy settings. In general, rate coding shows good results against all corruptions and is always more robust than the ANN baseline. Consequently, rate coding seems to be the best trade-off to deal with a large variety of image corruptions.

\begin{figure}[t]
     \centering
     \begin{subfigure}[b]{0.2\textwidth}
         \centering
         \includegraphics[width=\textwidth]{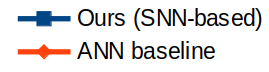}
         \caption{Legend}
         \label{fig:legendnoise_dv}
     \end{subfigure}
     \hfill
     \begin{subfigure}[b]{0.39\textwidth}
         \centering
         \includegraphics[width=\textwidth]{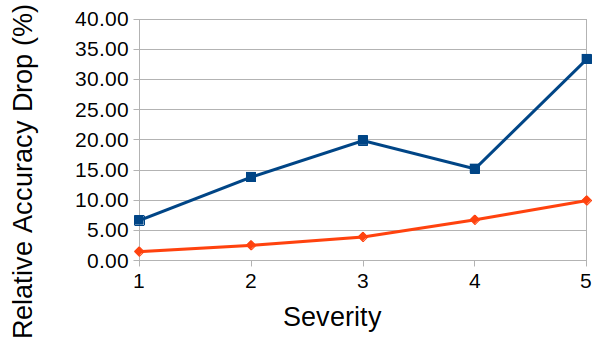}
         \caption{Hot Pixels}
         \label{fig:hotpixels_evolution}
     \end{subfigure}
     \hfill
     \begin{subfigure}[b]{0.39\textwidth}
         \centering
         \includegraphics[width=\textwidth]{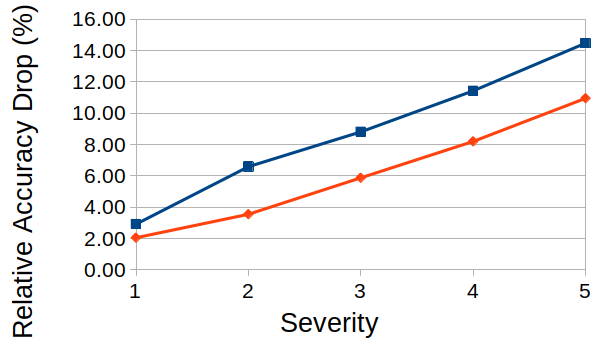}
         \caption{Background Activity Noise}
         \label{fig:ba_evolution}
     \end{subfigure}
        \caption{Evaluation of the relative accuracy drop score vs. the severity level of corruptions on event-based cameras}
        \label{fig:evolution_severity_dvs}
\end{figure}

\begin{table}[b]
\centering
\caption{Mean relative accuracy drop comparison on event-based inputs}
\label{tab:mean_acc_drop_dvs}
\begin{tabular}{c|c|c}
                      & \textbf{Hot pixels} & \textbf{Background activity} \\ \hline
\textbf{ANN baseline} & 4.92                & 6.13                         \\ \hline
\textbf{Ours}         & 17.78               & 8.85                        
\end{tabular}
\end{table}
As shown in Table \ref{tab:mean_acc_drop_dvs}, our model seems more sensitive to noises from event-based cameras compared to the ANN baseline. While both methods remain comparable for background activity noise, our SNN-based model performs poorly against hot pixels. Similarly, the evaluation of the relative accuracy drop score of our method grows faster than the ANN baseline for hot pixels noise, but the evaluation of this score for background activity noise is similar (shown in Figure \ref{fig:evolution_severity_dvs}). It can be explained by the sub-threshold dynamics of spiking neurons: they are constantly excited by the hot pixels events, which makes them spike more than expected and interfere with the feature extraction capacity of the convolutional SNN encoder. In comparison, hot pixels disturb the ANN baseline similarly to salt and pepper noise on the multiple event frames of $\mathbf{X}_T$, which is less harmful since artificial neurons do not integrate spikes from previous time-steps.

\subsection{Energy Efficiency Analysis}
\label{subsec:analysis_on_energy_consumption}

\begin{figure}[b]
\centering
\includegraphics[width=	0.48\textwidth]{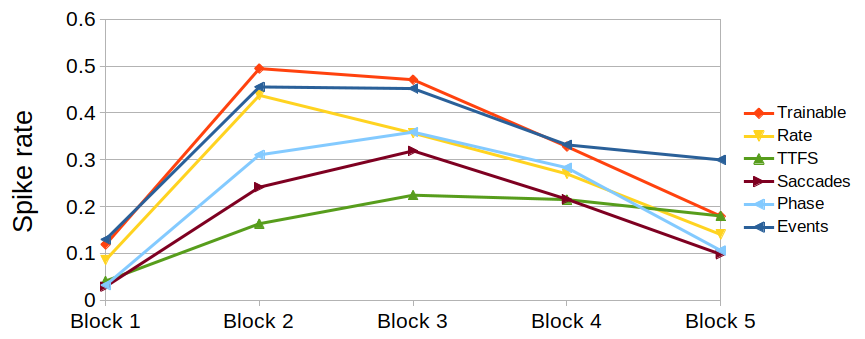}
\caption{Spike rate for each block of the SEW-ResNet-18 \cite{sew_resnet} architecture, depending on the neural coding scheme employed (or with event-based inputs)}
\label{fig:spike_rate}
\end{figure}

We provide an estimation of the energy consumption of our SNN-based model compared to ANNs, similarly to previous works \cite{snn_segmentation}. These estimations consider only the energy needed for Multiply And Accumulate (MAC) operations but do not take other factors into account (e.g., memory, peripheral circuit). To do so, we measure the rate of spikes for each layer of our model, since spiking neurons only consume energy when emitting a spike. The rate of spikes for a specific layer $l$ is given by:
\begin{equation}
Rs(l) = \frac{\text{\# spikes of } l \text{ over all time-steps}}{\text{\# neurons of } l}
\end{equation}
Since our model is a ResNet-like architecture \cite{resnet}, our network can be divided into 5 blocks, where each block processes feature maps with spatial resolution $2 \times$ smaller than the previous block. Figure \ref{fig:spike_rate} shows the spike rate of each block from our SEW-ResNet architecture (i.e., the average spike rate of all layers belonging to the residual block). Interestingly, the same pattern is observed for all coding schemes (or with event-based inputs): blocks 2, 3, and 4 have more activity compared to block 1 which receives the input spikes and block 5 which feeds the output accumulator module.

We can compute the total floating-point operations (FLOPs) of a layer $l$ of spiking neurons $FLOPs_{SNN}$ using the FLOPs of the same layer in an ANN $FLOPs_{ANN}$ (i.e., with non-spiking neurons) and the spike rate $Rs(l)$:
\begin{align}
\label{eq:flops_snn}
FLOPs_{SNN}(l) & = FLOPs_{ANN}(l) \times Rs(l) \\
\label{eq:flops_ann}
FLOPs_{ANN}(l) & = 
	\begin{cases}
	k^2 \times O^2 \times C_{in} \times C_{out} & \text{if } l \text{ is Convolutional} \\
    C_{in} \times C_{out} & \text{if } l \text{ is Linear.}
	\end{cases}
\end{align} In Equation \ref{eq:flops_ann}, $k$ is the kernel size, $O$ is the size of the output feature maps, $C_{in}$ is the number of input channels and $C_{out}$ is the number of output channels.

Using the total FLOPs across all layers, the total energy consumption of a model can be estimated on CMOS technology \cite{cmos}. Table \ref{tab:energy_operations} shows the energy cost of each relevant operation in a 45nm CMOS process. To perform a MAC operation, an ANN requires one addition (32bit FP ADD) and one FP multiplication (32bit FP MULT) \cite{mac_ann}. In comparison, SNNs only require one FP addition per MAC operation since they process binary spikes asynchronously. $E_{ANN}$ and $E_{SNN}$ denote the total energy consumption of ANNs and SNNs, respectively:
\begin{align}
E_{ANN} & = \sum\limits_{l}{FLOPs_{ANN}(l)} \times E_{MAC} \\
E_{SNN} & = \sum\limits_{l}{FLOPs_{SNN}(l)} \times E_{AC}
\end{align}

\begin{table}[b]
\centering
\caption{Energy table for a 45nm CMOS process (from \cite{snn_segmentation})}
\label{tab:energy_operations}
\begin{adjustbox}{max width=0.48\textwidth}
\begin{tabular}{cc}
\textbf{Operation} & \textbf{Energy} (pJ) \\ \hline
32bit FP MULT  ($E_{MULT}$)     & 3.7                  \\
32bit FP ADD  ($E_{ADD}$)     & 0.9                  \\
32bit FP MAC  ($E_{MAC}$)     & 4.6   ($= E_{MULT} + E_{ADD}$)             \\
32bit FP AC    ($E_{AC}$)    & 0.9                 
\end{tabular}
\end{adjustbox}
\end{table}

Finally, a comparison between $E_{ANN}$ and $E_{SNN}$ for every neural coding scheme and for event-based inputs is shown in Table \ref{tab:energy_consumption}. For the following experiments of this section, we use the models trained with $T=8$. In particular, the ratio $E_{ANN}/E_{SNN}$ describes the energy efficiency of our approach compared to the baseline ANN. Our method outperforms the ANN baseline by a large margin, being $44.82 \times$ to $126.6 \times$more efficient. As for the comparison between neural coding schemes, our results are consistent with previous works \cite{neural_coding} and with Table \ref{fig:spike_rate}: spike-intensive coding schemes (rate and phase coding) have higher spike rates and energy cost, while low-spike settings (TTFS and Saccades coding) are more efficient. The trainable coding scheme is shown to be the least efficient but is a particular case since the final representation strongly depends on the training step. By adding a spike penalization term \cite{stereospike} to the loss function, this coding scheme is expected to vary intensively from our results. On the other hand, the same efficiency as less effective coding schemes is observed for event-based inputs, which suggests that raw event-based sensors require a higher energy consumption for our object localization task than low-spike neural coding schemes.
 
\begin{table}[t]
\centering
\caption{Comparison of the energy consumption against a similar ANN architecture in the case of both frame-based and event-based contexts}
\label{tab:energy_consumption}
\begin{adjustbox}{max width=0.48\textwidth}
\begin{tabular}{c|ccc}
\textbf{}            & \textbf{$E_{ANN}$} (mJ)  & \textbf{$E_{SNN}$} (mJ) & \textbf{$E_{ANN} / E_{SNN}$} \\ \hline
\textbf{Trainable}   & \multirow{5}{*}{11129.44} & 248.34                   & 44.82                            \\
\textbf{Rate}        &                           & 208.6                    & 53.35                            \\
\textbf{TTFS}        &                           & 87.94                    & \textbf{126.6}                   \\
\textbf{Saccades}    &                           & {\ul114.69}                   & {\ul 97.04}                      \\
\textbf{Phase}       &                           & 141.79                   & 78.49                            \\ \hline
\textbf{Event-based} & 13 399.34                      & \textbf{294.63}                     & \textbf{45.48}                            
\end{tabular}
\end{adjustbox}
\end{table}
	
\subsection{Discussion on Coding Schemes}
From our various experiments on neural coding schemes, several conclusions can be drawn that differ from other previous studies \cite{falez_thesis,neural_coding} with SNNs and STDP learning rules. It may potentially show different behaviors of SNNs with surrogate gradient learning. Therefore, we expect it to set new guidelines on neural coding schemes with supervised SNNs on surrogate gradient. Four aspects are discussed: best accuracy, inference efficiency (i.e., performance of our networks for $T < 8$), overall robustness against image corruptions, and energy efficiency. We summarize our findings as follows. \textbf{Trainable Coding} is highly efficient in terms of accuracy (even with few time-steps) and robust, notably because it can be easily integrated with an SNN trained with backpropagation. However, achieving low energy consumption might require additional regularization during training \cite{regularization_low}. Our conclusions for \textbf{Rate Coding} and \textbf{TTFS Coding} differ importantly from previous studies \cite{falez_thesis,neural_coding} on STDP. Rate Coding is highly accurate, robust, and efficient on limited inference time-steps while TTFS Coding performs poorly in general but proves to be very robust. It shows that studies on bio-plausible learning rules may not generalize to other training strategies. Although having the poorest robustness by far, \textbf{Saccades Coding} shows to have low energy consumption and has comparable  performance on accuracy and on inference efficiency. \textbf{Phase Coding} shows similar characteristics to Rate Coding but has poorer performance in general, which highlights the fact that Rate Coding can be considered the best trade-off on the three aspects studied in our experiments.
\section{Conclusion}
\label{sec:conclusion}

In this work, we investigated the impacts of sensor noises and input coding schemes on single object localization, a vision task more sensitive to spatial information than classification, the most common  baseline in neuromorphic vision. We proposed a new SNN-based model for energy-efficient single object localization. We show that our novel approach can deal with both frame-based sensors (using various neural coding schemes) and event-based cameras using our adaptable output accumulator module. In addition, we evaluate the performance of our model on accuracy, inference efficiency, corruption robustness, and energy consumption and compare it to a similar ANN architecture. We report similar or better performance than ANNs in terms of accuracy and robustness and orders of magnitude better energy efficiency (up to $126.6 \times$). Finally, we summarize the pros and cons of popular neural coding schemes SNNs trained by surrogate gradient based on our experiments. As our observations differ importantly from previous works on bio-plausible learning rules \cite{neural_coding}, we believe that this summary can help researchers design more efficient SNN-based solutions trained with state-of-the-art backpropagation mechanisms.

\clearpage

\bibliographystyle{unsrt}  
\bibliography{egbib}  

\clearpage

\appendix
\section{Supplementary Materials}

\section{Neural Coding Schemes}
\begin{figure}[]
     \centering
     \begin{subfigure}[b]{0.99\textwidth}
         \centering
         \begin{tabular}{ccccccccc}
            \includegraphics[width=0.10\textwidth]{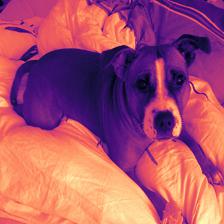} & \includegraphics[width=0.10\textwidth]{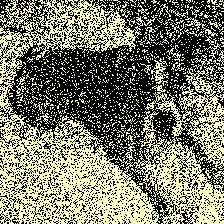} & \includegraphics[width=0.10\textwidth]{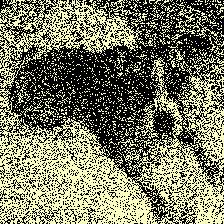} & \includegraphics[width=0.10\textwidth]{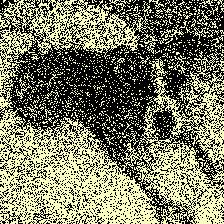} & \includegraphics[width=0.10\textwidth]{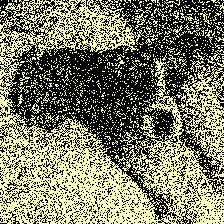} & \includegraphics[width=0.10\textwidth]{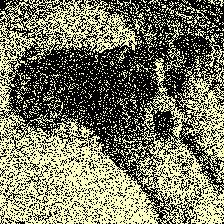} &
            \includegraphics[width=0.10\textwidth]{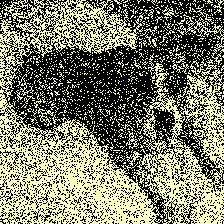} &
            \includegraphics[width=0.10\textwidth]{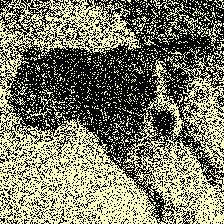} &
            \includegraphics[width=0.10\textwidth]{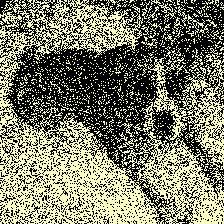} 
          \end{tabular}
         \caption{Rate Coding}
         \label{fig:ex_rate}
     \end{subfigure}
     \\
     \begin{subfigure}[b]{0.99\textwidth}
         \centering
         \begin{tabular}{ccccccccc}
            \includegraphics[width=0.10\textwidth]{noises/None_1.jpg} & \includegraphics[width=0.10\textwidth]{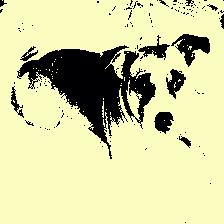} & \includegraphics[width=0.10\textwidth]{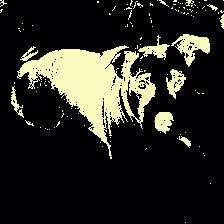} & \includegraphics[width=0.10\textwidth]{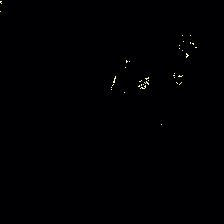} & \includegraphics[width=0.10\textwidth]{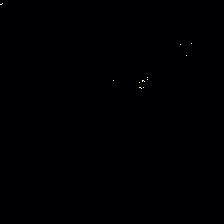} & \includegraphics[width=0.10\textwidth]{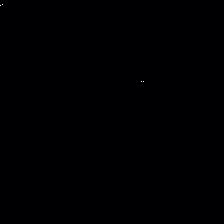} &
            \includegraphics[width=0.10\textwidth]{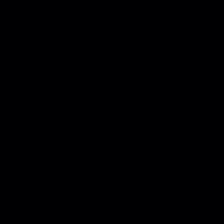} &
            \includegraphics[width=0.10\textwidth]{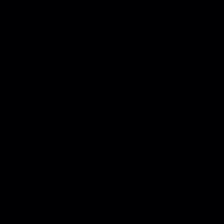} &
            \includegraphics[width=0.10\textwidth]{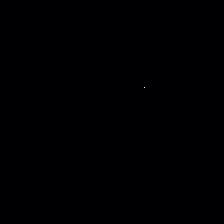} 
          \end{tabular}
         \caption{TTFS Coding}
         \label{fig:ex_ttfs}
     \end{subfigure}
     \\
     \begin{subfigure}[b]{0.99\textwidth}
         \centering
         \begin{tabular}{ccccccccc}
            \includegraphics[width=0.10\textwidth]{noises/None_1.jpg} & \includegraphics[width=0.10\textwidth]{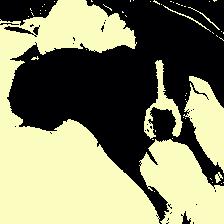} & \includegraphics[width=0.10\textwidth]{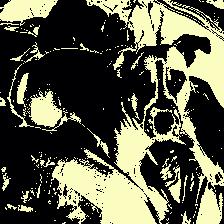} & \includegraphics[width=0.10\textwidth]{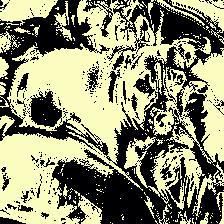} & \includegraphics[width=0.10\textwidth]{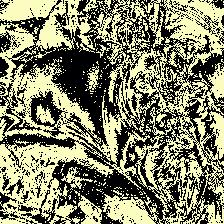} & \includegraphics[width=0.10\textwidth]{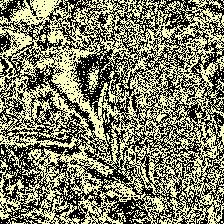} &
            \includegraphics[width=0.10\textwidth]{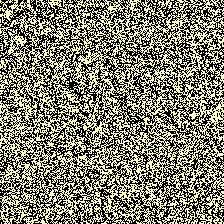} &
            \includegraphics[width=0.10\textwidth]{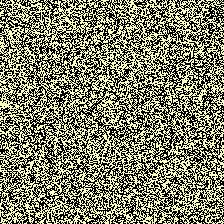} &
            \includegraphics[width=0.10\textwidth]{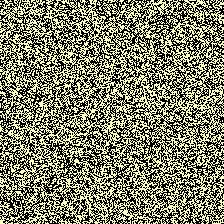} 
          \end{tabular}
         \caption{Phase Coding}
         \label{fig:ex_phase}
     \end{subfigure}
     \\
     \begin{subfigure}[b]{0.99\textwidth}
         \centering
         \begin{tabular}{ccccccccc}
            \includegraphics[width=0.10\textwidth]{noises/None_1.jpg} & \includegraphics[width=0.10\textwidth]{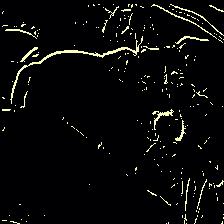} & \includegraphics[width=0.10\textwidth]{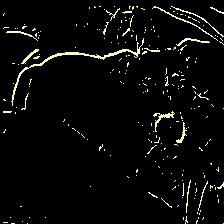} & \includegraphics[width=0.10\textwidth]{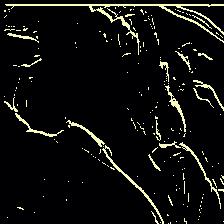} & \includegraphics[width=0.10\textwidth]{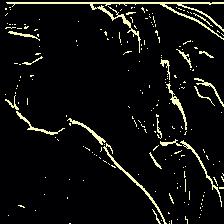} & \includegraphics[width=0.10\textwidth]{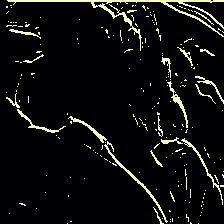} &
            \includegraphics[width=0.10\textwidth]{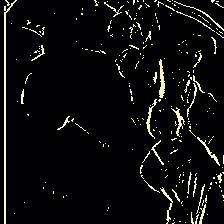} &
            \includegraphics[width=0.10\textwidth]{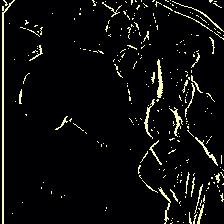} &
            \includegraphics[width=0.10\textwidth]{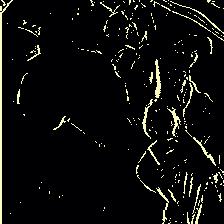} 
          \end{tabular}
         \caption{Saccades Coding}
         \label{fig:ex_saccades}
         \end{subfigure}
         \\
         \begin{subfigure}{0.99\textwidth}
         \hfill
             \includegraphics[width=0.87\textwidth]{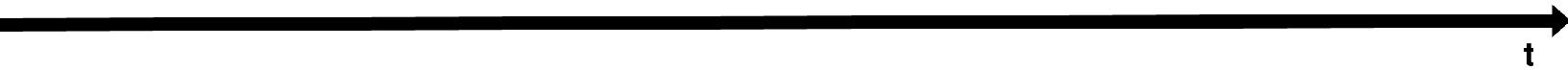}
         \end{subfigure}
        \caption{Examples for each neural coding scheme}
        \label{fig:nc_examples}
\end{figure}

In this section, we further describe the various neural coding schemes used to convert a static image $\mathbf{I} \in \mathbb{R}^{C \times H \times W}$ into spike trains and detail their implementations. In addition, examples of the resulting $\mathbf{X}_T$ are shown in Figure \ref{fig:nc_examples} for each coding scheme, with a defined total number of time-steps $T = 8$. Trainable Coding is not illustrated because of too high number of channels (i.e., $C = 32$ after neural coding).

\subsection{Rate Coding}
Our rate coding definition and implementation is from the SnnTorch framework \cite{snntorch}. Each pixel value $I_{cij}$ (normalized between 0 and 1) from the static image $\mathbf{I}$ represents the probability a spike occurs at any time-step. It is treated as a Bernouilli trial $B(n,p)$, where the number of trials is $n=1$, and $p = I_{cij}$ is the probability of success (i.e., a spike occurs).

For exemple, white pixels ($I_{cij} = 1$) represent a probability of 100\% of spiking, while black pixels ($I_{cij} = 0$) will never spike.

\subsection{Time-To-First-Spike (TTFS) Coding}
We use the implementation and definition of TTFS Coding from SnnTorch \cite{snntorch}.

The objective is to obtain a precise spike timing (representing the time where the unique spike occurs) from a normalized pixel value. To do so, the logarithmic dependence between an input intensity value and the related spike timing is derived using an RC circuit model, where the normalized value of the input pixel $I_{cij}$ is represented as a constant current injection (see details in Appendix B.2 of \cite{snntorch}). For short, the output spike timing $t(I_{cij})$ of an input pixel is given by:
\begin{align}
t(I_{cij}) = \begin{cases}
      \tau  \Big[\ln(\frac{I_{cij}}{I_{cij} - 0.01})\Big], & \text{if}\ I_{cij} > 0.01 \\
      T, & \text{otherwise}
    \end{cases}
\end{align}
where $\tau = 1$ is the time-constant of the RC circuit model.

\subsection{Phase Coding}
The objective of Phase Coding is to decompose the total duration allocated for the spike train in multiple phases. These multiple phases originally represent generated oscillation rythms that have been experimentally observed in the hippocampus and olfactory system \cite{biology_phase}.

In our work, we use a simple and effective implementation of phase coding based on the 8-bit representation of unnormalized input pixels (i.e., values from 0 to 255) \cite{phase_coding}. Consequently, a total cycle of phases consists of 8 time-steps. Then, these $8$ phases are successively repeated/discarded until the $T$ time-steps are completed. The weights $w_s(t) = 2^{-1+mod(t-1,8)}$ used to replicate the significance of each bit in the 8-bit representation can be seen as synaptic weights directly applied to the spike tensor $\mathbf{X}_T$.

Even though this Phase Coding process is not bio-plausible, it is considered a good strategy to implement phase coding in neuromorphic solutions \cite{neural_coding}.

\subsection{Saccades Coding}
\begin{figure}[]
\centering
\includegraphics[width=0.90\textwidth]{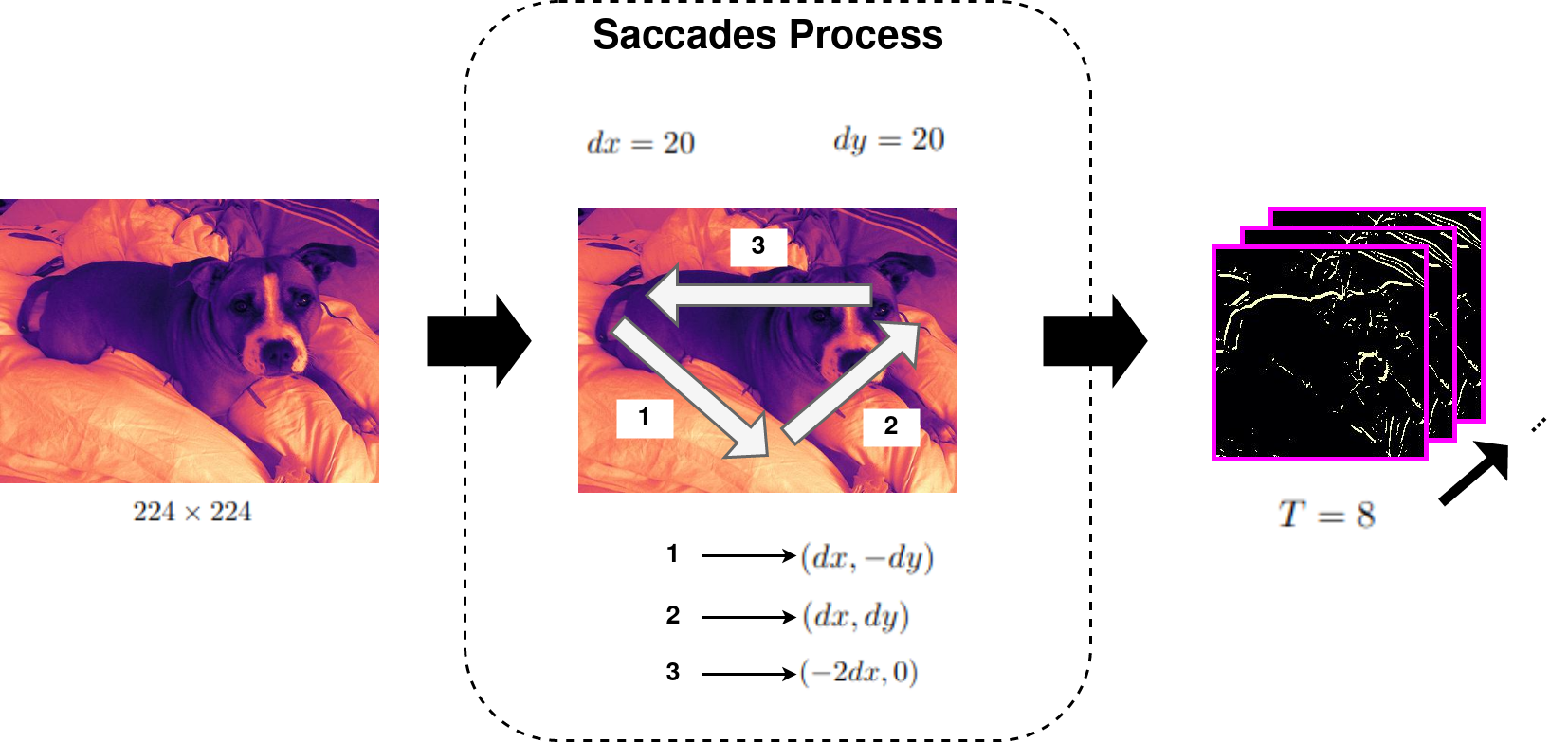}
\caption{Overview of the Saccades Coding process. Three translations are successively applied in the order shown}
\label{fig:saccade_overview}
\end{figure}

Our implementation of Saccades Coding simulates the same procedure from N-MNIST \cite{nmnist}, but only digitally (whereas N-MNIST uses a physical event camera on a motorized pan-tilt). That is,  from the original image $\mathbf{I}$, we create a sequence of $T$ frames that represent three translations (the ``saccades''). These translations are used to progressively move the static image based on two distance values $dx$ and $dy$ (in pixels). Then, a delta modulation process is applied to this sequence of $T$ frames to obtain the final spike tensor $\mathbf{X}_T$. We use the implementation of the delta modulation process from SnnTorch \cite{snntorch}, with a threshold of $0.1$. Figure \ref{fig:saccade_overview} shows an overview of Saccades Coding and especially describe the three translations. In Figure \ref{fig:ex_saccades}, we observe the effects of these saccades that look similar to samples from \cite{nmnist} (a comparable example is shown from N-Caltech101 in Figure \ref{fig:dvs_examples}).

\subsection{Trainable Coding}
In Trainable Coding, the first block convolutional layer of our SEW-ResNet-18 \cite{sew_resnet} is treated as the neural coding scheme, i.e., the image $\mathbf{I}$ is directly fed into the layer. The output of this layer has a resolution of $32 \times \frac{H}{2} \times \frac{W}{2}$. Then, an Integrate-and-Fire activation is applied and the resulting spikes are repeated over the $T$ time-steps, which gives the resulting $\mathbf{X}_T$. Consequently, this coding highly discriminates the low-level features introduced in the network, but can be optimized end-to-end using backpropagation.

\section{Image Corruptions}
In this section, we describe the corruptions used for both modalities (event-based and frame-based inputs) and detail the implication of a severity level for all corruptions.

The objective of the severity level is to obtain a corruption more (or less) destructive so that the robustness of our different models can be estimated. For a given corruption, the severity level defines one or more parameter values that result in more (or less) corrupted inputs.

\subsection{Corruption of Static Images}
\begin{figure}[]
     \centering
     \begin{subfigure}[b]{0.99\textwidth}
         \centering
         \begin{tabular}{cccccc}
            \includegraphics[width=0.16\textwidth]{noises/None_1.jpg} & \includegraphics[width=0.16\textwidth]{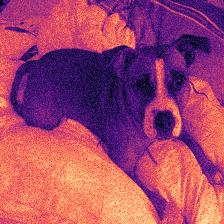} & \includegraphics[width=0.16\textwidth]{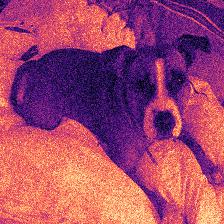} & \includegraphics[width=0.16\textwidth]{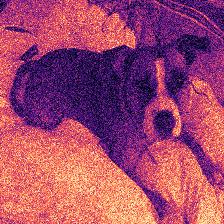} & \includegraphics[width=0.16\textwidth]{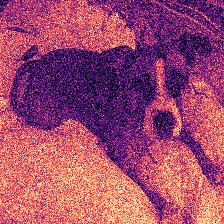} & \includegraphics[width=0.16\textwidth]{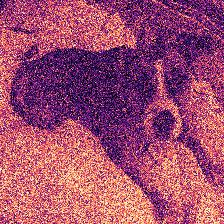}
          \end{tabular}
         \caption{Gaussian Noise}
         \label{fig:ex_gauss}
     \end{subfigure}
     \\
     \begin{subfigure}[b]{0.99\textwidth}
         \centering
         \begin{tabular}{cccccc}
            \includegraphics[width=0.16\textwidth]{noises/None_1.jpg} & \includegraphics[width=0.16\textwidth]{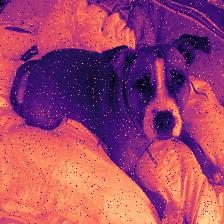} & \includegraphics[width=0.16\textwidth]{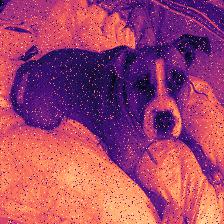} & \includegraphics[width=0.16\textwidth]{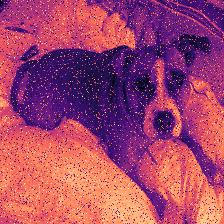} & \includegraphics[width=0.16\textwidth]{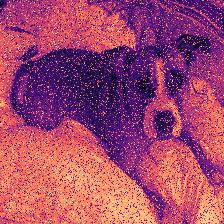} & \includegraphics[width=0.16\textwidth]{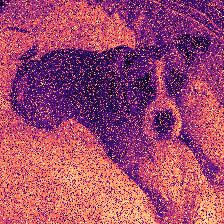}
          \end{tabular}
         \caption{Salt \& Pepper Noise}
         \label{fig:ex_sp}
     \end{subfigure}
     \\
     \begin{subfigure}[b]{0.99\textwidth}
         \centering
         \begin{tabular}{cccccc}
            \includegraphics[width=0.16\textwidth]{noises/None_1.jpg} & \includegraphics[width=0.16\textwidth]{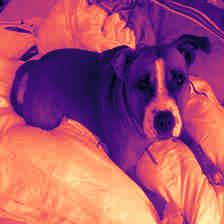} & \includegraphics[width=0.16\textwidth]{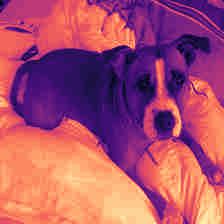} & \includegraphics[width=0.16\textwidth]{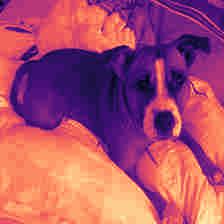} & \includegraphics[width=0.16\textwidth]{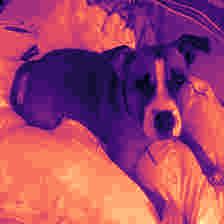} & \includegraphics[width=0.16\textwidth]{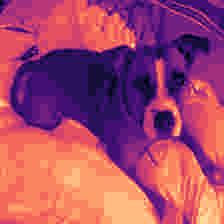}
          \end{tabular}
         \caption{JPEG Compression}
         \label{fig:ex_jpeg}
     \end{subfigure}
     \\
     \begin{subfigure}[b]{0.99\textwidth}
         \centering
         \begin{tabular}{cccccc}
            \includegraphics[width=0.16\textwidth]{noises/None_1.jpg} & \includegraphics[width=0.16\textwidth]{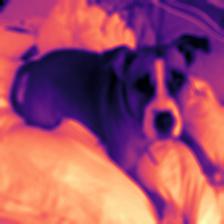} & \includegraphics[width=0.16\textwidth]{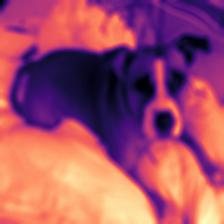} & \includegraphics[width=0.16\textwidth]{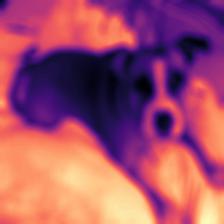} & \includegraphics[width=0.16\textwidth]{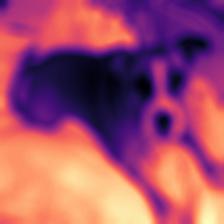} & \includegraphics[width=0.16\textwidth]{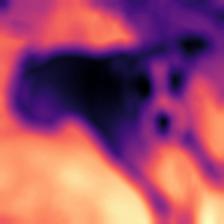}
          \end{tabular}
         \caption{Defocus Blur}
         \label{fig:ex_defocus}
         \end{subfigure}
    \\
    \begin{subfigure}[b]{0.99\textwidth}
         \centering
         \begin{tabular}{cccccc}
            \includegraphics[width=0.16\textwidth]{noises/None_1.jpg} & \includegraphics[width=0.16\textwidth]{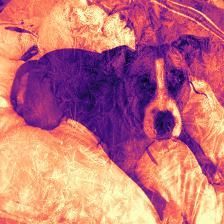} & \includegraphics[width=0.16\textwidth]{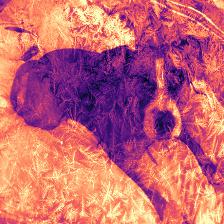} & \includegraphics[width=0.16\textwidth]{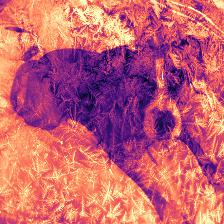} & \includegraphics[width=0.16\textwidth]{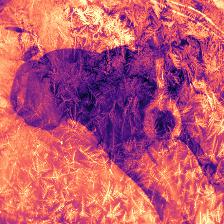} & \includegraphics[width=0.16\textwidth]{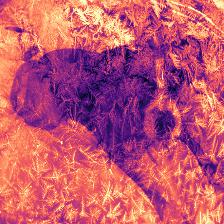}
          \end{tabular}
         \caption{Frost Perturbations}
         \label{fig:ex_frost}
     \end{subfigure}
        \caption{Examples of corruptions. First column shows a clean example. The growing levels of severity (from $1$ to $5$) are shown in the remaining columns (from left to right)}
        \label{fig:corr_example}
\end{figure}

The corruptions used for static images are from \cite{benchmark_corruptions} and we reuse the original implementation\footnote{\url{https://github.com/hendrycks/robustness}}. Table \ref{tab:param_static} summarizes the parameter values defined by the severity level for each corruption. Examples of corruptions (with a growing level of severity) are given in Figure \ref{fig:corr_example}.

\begin{table}[t]
\centering
\caption{Parameters for all corruptions used in our experiments on static images}
\label{tab:param_static}
\begin{tabular}{c|c|ccccc}
\multirow{2}{*}{\textbf{Noise}} & \multirow{2}{*}{\textbf{Parameter(s)}} & \multicolumn{5}{c}{\textbf{Severity Level}}                      \\
                                &                                        & \textbf{1} & \textbf{2} & \textbf{3} & \textbf{4}  & \textbf{5}  \\ \hline
Gaussian Noise                  & $\sigma$                                  & 0.08       & 0.12       & 0.18       & 0.26        & 0.38        \\
Salt \& Pepper                   & $sp$                                     & 0.03       & 0.06       & 0.09       & 0.17        & 0.27        \\
JPEG Compression                & Quality \%                              & 25         & 18         & 15         & 10          & 7           \\
Defocus                         & $r$                                      & 3          & 4          & 6          & 8           & 10          \\
Frost                           & $(\rho, \omega)$                           & (1, 0.4)   & (0.8, 0.6) & (0.7, 0.7) & (0.65, 0.7) & (0.6, 0.75)
\end{tabular}
\end{table}

\subsubsection{Gaussian Noise}
We add a gaussian noise $(0, \sigma)$ to the original input image, i.e. a normal distribution is added to the original input image. The severity level corresponds to the standard deviation $\sigma$ of the normal distribution.

\subsubsection{Salt \& Pepper Noise}
A certain proportion $sp$ of pixels are randomly set to $0$ of $1$ values. The severity level represents the proportion $sp$ of pixels corrupted in the original image.

\subsubsection{JPEG Compression}
A JPEG compression algorithm is applied to an input image, with a defined quality percentage. The severity level corresponds to this quality percentage.

\subsubsection{Defocus Blur}
To simulate a situation where the focal plane of a camera is away from the sensor plane, a common strategy \cite{outoffocus} is to apply an aliased disk kernel of radius $r$ that is the parameter defined by the severity level (i.e., a larger $r$ increases the intensity of defocus blur).

\subsubsection{Frost Perturbation}
\begin{figure}[]
\centering
\includegraphics[width=0.50\textwidth]{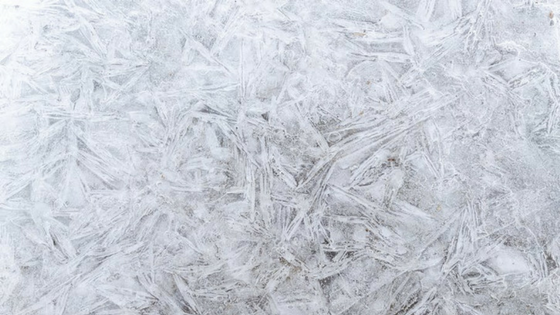}
\caption{Example of ice crystal image used in the Frost Perturbation corruption}
\label{fig:frost_ex}
\end{figure}

To efficiently simulate frost perturbations, samples of ice crystals are randomly selected from pre-generated images and added to the original input image $\mathbf{I}$. An example of pre-generated images is shown in Figure \ref{fig:frost_ex}. We denote $\mathbf{I}_{frost} \in \mathbb{R}^{C\times H \times W}$ as the selected sample of ice crystals.

The severity level defines how much the ice crystals occlude the original image. To do so, we define two parameters $(\rho,\omega)$ that correspond to the intensity (between 0 and 1) of $\mathbf{I}$ and $\mathbf{I}_{frost}$, respectively. Consequently, the resulting corrupted image is obtained from $\rho \mathbf{I} + \omega \mathbf{I}_{frost}$.

\subsection{Corruption of Event-based Inputs}
\begin{figure}[]
     \centering
     \begin{subfigure}[b]{0.99\textwidth}
         \centering
         \includegraphics[width=\textwidth]{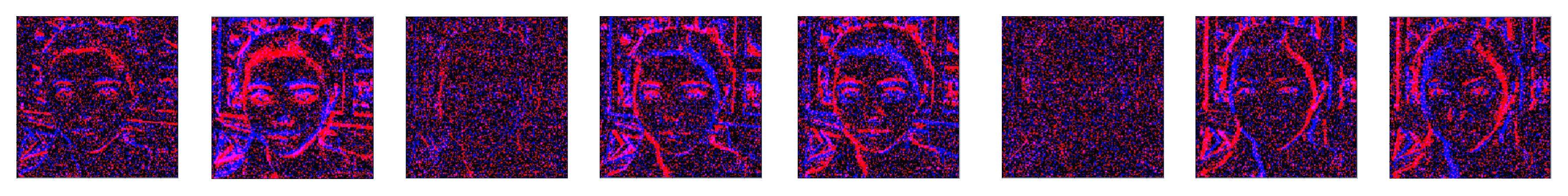}
         \caption{Example of Hot Pixels noise}
         \label{fig:ex_hotpix}
     \end{subfigure}
     \\
     \begin{subfigure}[b]{0.99\textwidth}
         \centering
         \includegraphics[width=\textwidth]{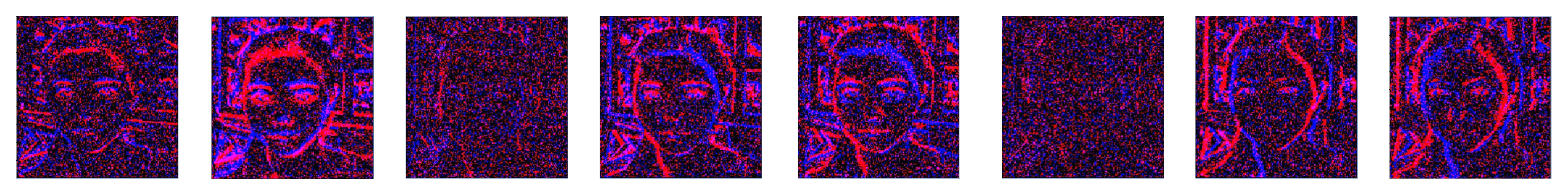}
         \caption{Example of Background Activity noise}
         \label{fig:ex_baa}
     \end{subfigure}
     \\
     \begin{subfigure}[b]{0.99\textwidth}
         \centering
         \includegraphics[width=\textwidth]{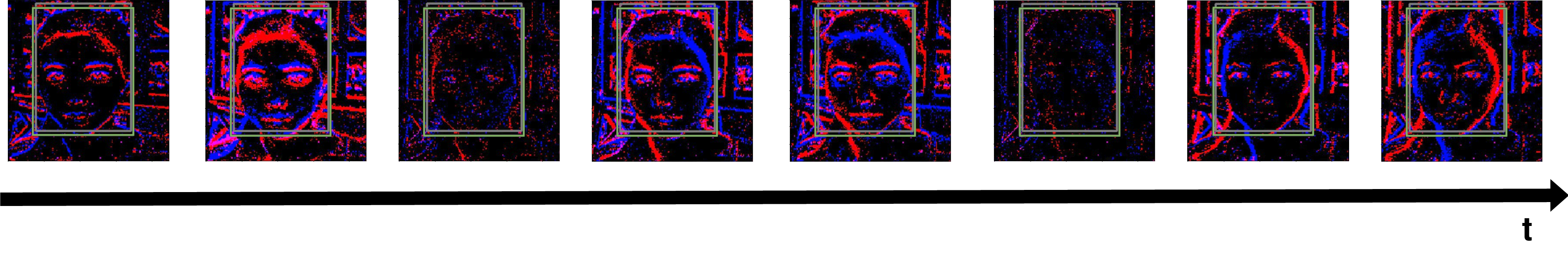}
         \caption{Clear sample, with example of prediction. The predicted bounding box is shown in grey, the ground truth is shown in green}
         \label{fig:ex_ex_dvs}
     \end{subfigure}
        \caption{An example of event-based input from N-Caltech101 \cite{nmnist}, with $T = 8$}
        \label{fig:dvs_examples}
\end{figure}

The corruptions for event-based inputs (i.e., hot pixels and background activity noises) are directly implemented on the discretized spike tensor $\mathbf{X}_T$. Table \ref{tab:param_event} details the parameter values that depend on the severity level and Figure \ref{fig:dvs_examples} illustrates the resulting corrupted samples for $T = 8$.

\begin{table}[t]
\centering
\caption{Values of parameters for all corruptions used in our experiments on event-based inputs}
\label{tab:param_event}
\begin{tabular}{c|c|ccccc}
\multirow{2}{*}{\textbf{Noise}} & \multirow{2}{*}{\textbf{Parameter(s)}}                                & \multicolumn{5}{c}{\textbf{Severity Level}}                    \\
                                &                                                                       & \textbf{1} & \textbf{2} & \textbf{3} & \textbf{4} & \textbf{5} \\ \hline
Hot Pixels                      & \begin{tabular}[c]{@{}c@{}}Hot Pixels\\ Proportion (\%)\end{tabular} & 0.03       & 0.06       & 0.09       & 0.17       & 0.27       \\
Background Activity             & $\lambda$                                                               & 0.08       & 0.12       & 0.18       & 0.26       & 0.38      
\end{tabular}
\end{table}

\subsubsection{Hot Pixels}
As hot pixels are pixels that fire constantly (e.g. due to faulty hardware), it can be efficiently simulated by randomly selecting a set of pixel coordinates and fixing their values to $1$ (i.e., spiking), which corresponds to applying the same function to the spike tensor $\mathbf{X}_T$ at every time-step. The level of severity defines the percentage of pixel coordinates from the original input that are treated as hot pixels.

\subsubsection{Background Activity}
Since \cite{background_activity} has shown that background activity noise can be simulated by a time-independent Poisson process, we employ the same strategy in our work. At each time-step, we draw samples from a Poisson distribution with a parameter $\lambda$ being the expected number of events occuring in the time interval (see the documentation of Numpy \cite{numpy} for a random Poisson distribution, where $\lambda$ is the parameter \textit{lam}). The level of severity corresponds to the parameter $\lambda$, which influences the number of events corrupted by the Poisson noise.

\end{document}